\documentclass{article}

\usepackage[ final]{neurips_2025}

\usepackage{tikz}
\usetikzlibrary{tikzmark}

\def\hlinewd#1{\noalign{\ifnum0=`}\fi\hrule \@height #1 \futurelet \reserved@a\@xhline}

\definecolor{brown1}{RGB}{227, 204, 194}  % 定义自定义蓝色（钢蓝色）
\definecolor{brown2}{RGB}{247, 219, 189}  % 定义自定义蓝色（钢蓝色）

\definecolor{myblue}{RGB}{42, 116, 174}  % 定义自定义蓝色（钢蓝色）
\definecolor{fullmarkcolor}{RGB}{214, 233, 213} % Light green for Full mark row
\definecolor{opensource}{RGB}{255, 243, 206}
\definecolor{opensource2}{RGB}{217, 231, 252}
\definecolor{streaming}{RGB}{214, 239, 244} % Light red for W/O videos row
\definecolor{closesource}{RGB}{184, 219, 179}
\definecolor{blue1}{rgb}{0.8, 0.9, 1.0}  % 浅蓝色
\definecolor{blue2}{rgb}{0.7, 0.85, 1.0}  % 中蓝色
\definecolor{blue3}{rgb}{0.4, 0.7, 0.9}  % 深蓝色

\definecolor{scifiBlue}{rgb}{0.2, 0.7, 1.0}   % 明亮的电子蓝
\definecolor{scifiRed}{rgb}{1.0, 0.2, 0.2}     % 炽热的科幻红
\definecolor{scifiGreen}{rgb}{0.4, 1.0, 0.3} % 备选：荧光绿
\definecolor{scifiPurple}{rgb}{0.6, 0.2, 0.8} % 备选：深邃的紫罗兰
\definecolor{demphcolor}{gray}{.5}

\usepackage{pifont}
\usepackage{enumitem}
\usepackage{adjustbox}
\usepackage[utf8]{inputenc} % allow utf-8 input
\usepackage[T1]{fontenc}    % use 8-bit T1 fonts
\usepackage{hyperref}       % hyperlinks
\usepackage{url}            % simple URL typesetting
\usepackage{booktabs}       % professional-quality tables
\usepackage{amsfonts}       % blackboard math symbols
\usepackage{nicefrac}       % compact symbols for 1/2, etc.
\usepackage{microtype}      % microtypography
\usepackage{xcolor}         % colors
\usepackage{colortbl} 
\usepackage{graphicx}
\usepackage{amsmath}
\usepackage{multirow} 
\usepackage{caption}
\usepackage{mathtools}

\definecolor{lightgold}{RGB}{255, 242, 217}

\usepackage{inconsolata} % Or any other monospaced font package
\usepackage{tcolorbox} % For creating nice boxes
\tcbuselibrary{skins}
\tcbset{
    colback=gray!5,
    colframe=black!70,
    fonttitle=\bfseries,
    sharp corners,
    boxrule=0.5pt,
    verbatim % <--- This line is important
}
\usepackage{array} 

\definecolor{miragepink}{RGB}{255, 0, 127} % 定义一个名为 'miragepink' 的新颜色

\hypersetup{
    colorlinks=true, % 启用颜色链接
    linkcolor=miragepink, % 设置内部链接（如目录、图表引用）的颜色
    citecolor=miragepink, % 设置参考文献引用的颜色
    urlcolor=miragepink, % 设置 URL 链接的颜色
}

\title{ Eyes Wide Open: \\
Ego Proactive Video-LLM for Streaming Video}

\makeatletter
\def\blfootnote{\xdef\@thefnmark{}\@footnotetext}
\makeatother

\author{
Yulin Zhang$^{1}$ \hspace{1.5em} Cheng Shi$^{3}$  \hspace{1.5em} Yang Wang$^{1}$ \hspace{1.5em} Sibei Yang$^{2\dagger}$\\
{$^1$ShanghaiTech University}\hspace{1.5em}
{$^2$School of Computer Science and Engineering, Sun Yat-sen University} \\ {$^3$School of Computing and Data Science, The University of Hong Kong} \\
\textbf{Project Page: }\url{https://zhangyl4.github.io/publications/eyes-wide-open/}
}

\begin{document}
% 在 \author 环境之后，\maketitle 之前（或您希望脚注出现的位置）添加：
\blfootnote{$^\dagger$Corresponding author is Sibei Yang.}
\maketitle

\begin{abstract}
Envision an AI capable of functioning in human-like settings, moving beyond mere observation to actively understand, anticipate, and proactively respond to unfolding events. Towards this vision, we focus on the innovative task where, \textit{given ego-streaming video input, an assistant proactively answers diverse, evolving questions at the opportune moment, while maintaining synchronized perception and reasoning.} This task embodies three key properties: (1) Proactive Coherence, (2) Just-in-Time Responsiveness, and (3) Synchronized Efficiency.
To evaluate and address these properties, we first introduce ESTP-Bench (Ego Streaming Proactive Benchmark) alongside the ESTP-F1 metric—a novel framework designed for their rigorous assessment. Secondly, we propose a comprehensive technical pipeline to enable models to tackle this challenging task. This pipeline comprises: (1) a data engine, (2) a multi-stage training strategy, and (3) a proactive dynamic compression technique. Our proposed model effectively addresses these critical properties while outperforming multiple baselines across diverse online and offline benchmarks.
\end{abstract}    
\section{Introduction}
\label{sec:intro}

Imagine an AI assistant that follows you through your day—assembling furniture, searching for misplaced keys~\cite{barmannWhereDidLeave2022, golettoAMEGOActiveMemory2024}, or preparing a meal~\cite{tangCOINLargescaleDataset2019, song2023egod}—not just watching, but understanding, anticipating~\cite{yangEgoLifeEgocentricLife2025}, and responding proactively when needed as events unfold. 
To function in such human-like settings, where visual input is egocentric and continuously streaming, and user needs shift from moment to moment, the assistant must go beyond passive observation. It should be able to interpret the present, anticipate what comes next, and respond at exactly the right moment, all in real time. 

\begin{figure}[t]
    \centering
    \includegraphics[width=1.0\linewidth]{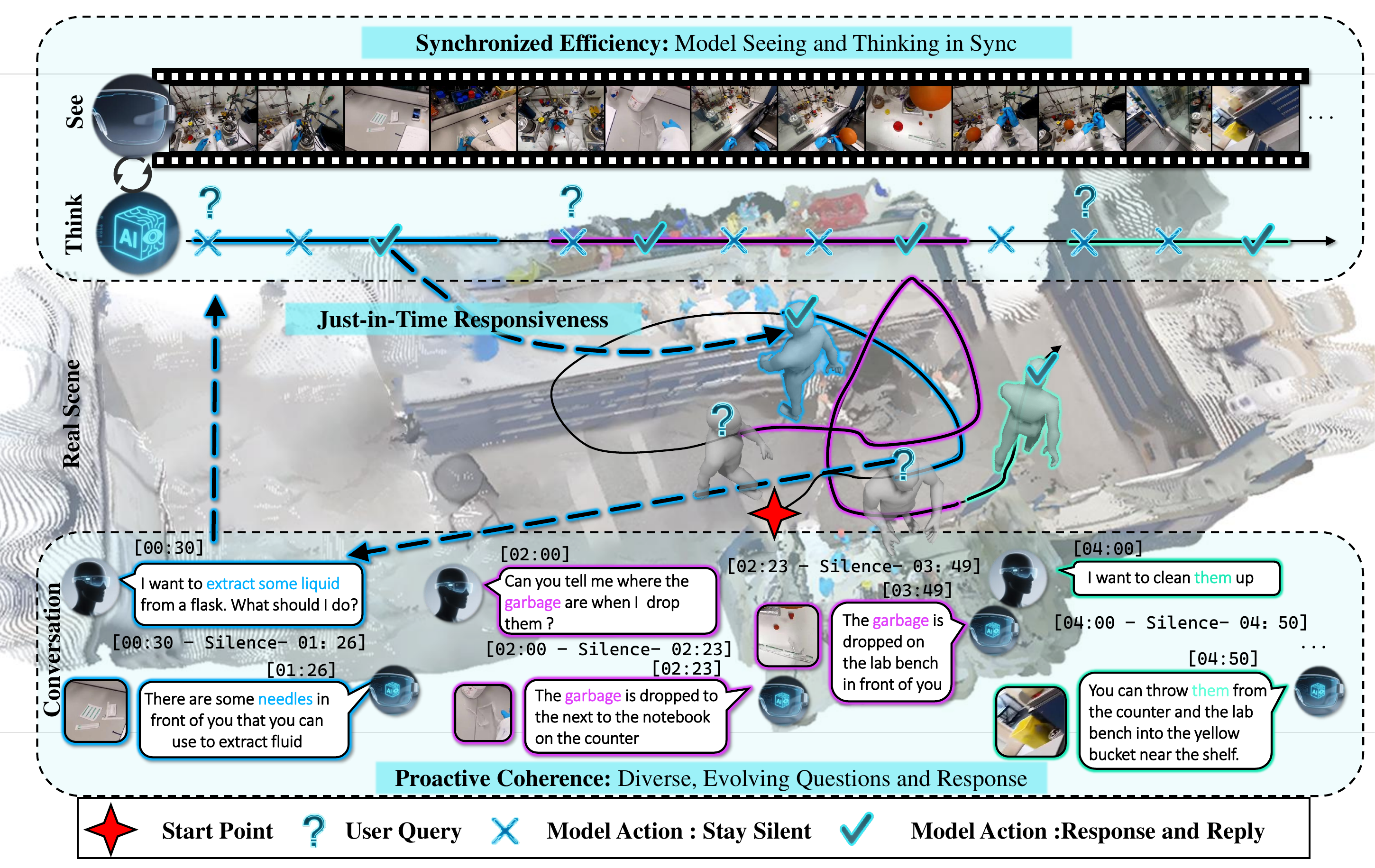}
    \caption{\textbf{An illustrative example of the ESTP task.} The figure is structured in three layers: the top layer depicts the model's continuous visual processing and decision-making (\textit{See and Think}), the middle layer shows the real-world egocentric scene with the human's trajectory, and the bottom layer presents the human-model conversation.}
    \label{fig:teaser}
    \vspace{-0.2cm}
\end{figure}

As a first step toward this vision, we narrow our focus to perception and understanding in egocentric streaming video, with a particular emphasis on the following innovative task: 
\textcolor{scifiGreen!70!black}{\textit{\textbf{Given ego-streaming video input, the assistant proactively answers to diverse and evolving questions at the right moment, while seeing and thinking in sync,}}} as shown in Fig.~\ref{fig:teaser}. This task relies on three key properties:
\begin{itemize}[leftmargin=*, itemsep=0pt]
    \item Proactive Coherence: handling diverse question types, responding even when answers depend on future visual streams (proactivity), and maintaining contextual consistency across related questions. In ego-streaming scenarios, questions often go beyond the current frame, referencing future events or past observations. As shown in Fig.~\ref{fig:teaser}, the segment of the conversation highlighted in green is contextually dependent on the content within the segment highlighted in purple. Such queries require temporal integration of past and present information, followed by proactive answering as relevant visual evidence emerges. 
    \item Just-in-Time Responsiveness: determining when to answer based on visual readiness, neither too soon nor too late, and only when necessary. Responding before enough evidence is available can lead to mistakes, while answering too late may miss the opportunity to help. Equally important is staying silent when uncertain and avoiding unnecessary repetition. As shown in the blue-highlighted segment of Fig.~\ref{fig:teaser}, it is necessary to remain silent until the ``face to counter''. The assistant must continuously track the evolving visual context and respond at the earliest reliable moment. 
    \item Synchronized Efficiency: ensuring that answering and visual perception proceed in sync without delay. Responses should not come at the cost of missing new visual input; perception and reasoning must remain temporally aligned. Regarding the purple segment depicted in Fig.~\ref{fig:teaser}, maintaining synchronization is crucial to prevent missed answers. This requires answering while continuously observing, with zero latency, while also ensuring time and memory efficiency as the number of incoming frames grows over time. 
\end{itemize}

\begin{figure}[t]
    \centering
    \begin{minipage}[t]{0.28\textwidth} % Adjusted width, using [t] for top alignment
        \centering
        \vspace{0pt}
        \includegraphics[width=0.94\linewidth]{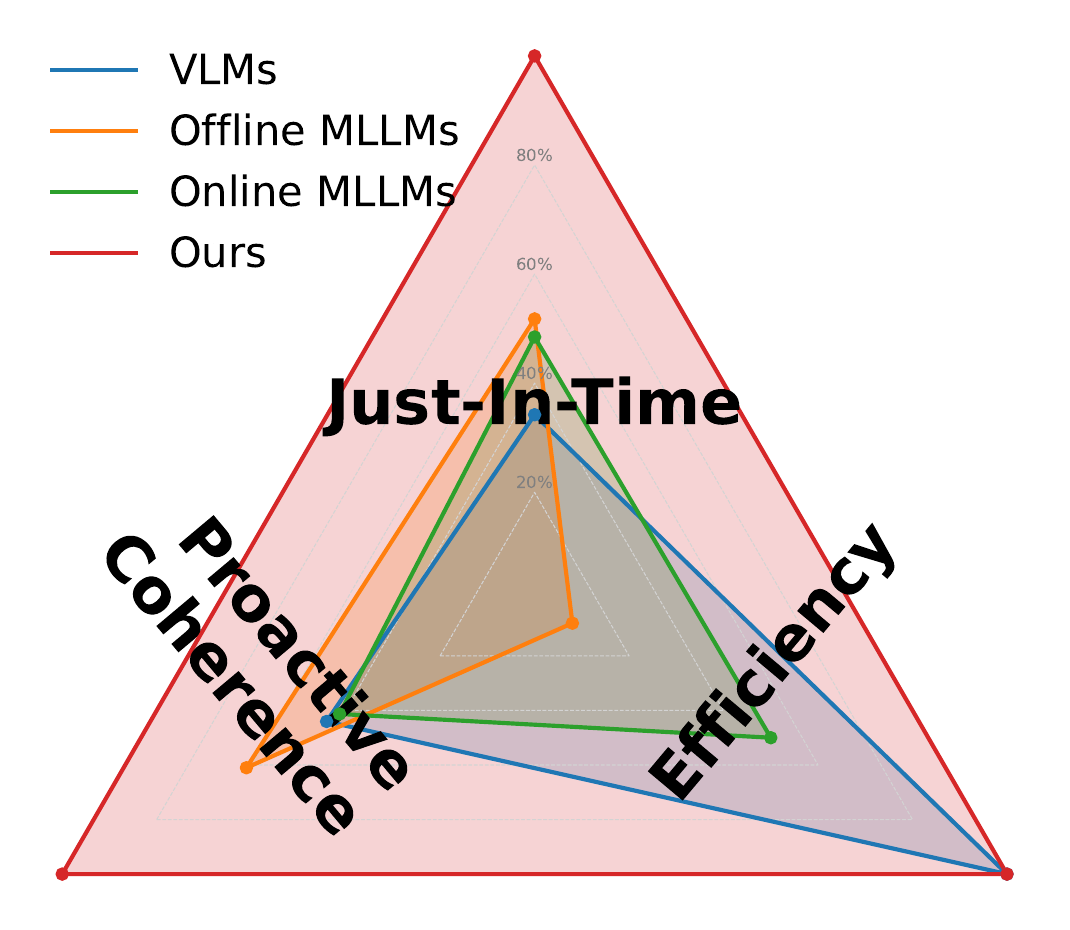}
        
       \caption{ESTP Triangle of Impossibility shows trade-offs among the three dimensions: Proactive Coherence, Just-in-Time responsiveness, and Efficiency, which are quantified by contextual performance, recall, and FPS.}
        \label{fig:t2}
    \end{minipage}%
    % \hfill % Adds flexible horizontal space
    \begin{minipage}[t]{0.71\textwidth} % Adjusted width
        \centering
        \vspace{0pt}
        \resizebox{\textwidth}{!}{ % 强制表格适应页面宽度
    % 增加了第 6 列作为空列
    \begin{tabular}{l c c c c c c c c c } 
    % \large
    % 10 列: Dataset (1), Ques. Type (2), Proactive (3-5), EMPTY (6), Evaluation (7-9), # Ques. (10)
        \specialrule{1pt}{0pt}{0pt} % 粗顶线
        % 第一层表头
        \multirow{2}{*}{Dataset} & \multirow{2}{*}{Ques. Type} & \multicolumn{3}{c}{Proactive Type} & \multicolumn{1}{c}{}% 新增的空列表头
        & \multicolumn{3}{c}{JIT Responsiveness Eval.} & \multirow{2}{*}{\# Ques.} \\
        % \cline{3-5}\cline{6-8} % 原分隔线，需要调整跨列范围
        \cline{3-5}\cline{7-9} % 调整后的分隔线范围
        % 第二层表头 (子表头)
        & & Exp. & Imp. & Cont. &% 新增的空列
        & Ans. Turn & Is Prec. & Timeliness & \\
        \hline % 表头和数据之间的细分隔线
        \textbf{\textit{ Online Benchmark}} \\
        VStream ~\cite{zhangFlashVStreamMemoryBasedRealTime2024} & OE & \textcolor{scifiPurple}{\ding{56}} & \textcolor{scifiPurple}{\ding{56}} & \textcolor{scifiPurple}{\ding{56}} & & S & \textcolor{scifiPurple}{\ding{56}} & \textcolor{scifiPurple}{\ding{56}} & 3,500 \\
        StreamingBench~\cite{linStreamingBenchAssessingGap2024} & MC & \textcolor{scifiPurple}{\ding{56}} & \textcolor{scifiPurple}{\ding{56}} & \textcolor{scifiPurple}{\ding{56}} & & S & \textcolor{scifiPurple}{\ding{56}} & \textcolor{scifiPurple}{\ding{56}} & 4,500 \\
        StreamingBench (PO)~\cite{linStreamingBenchAssessingGap2024} & Q-Match & \textcolor{scifiBlue}{\ding{52}} & \textcolor{scifiPurple}{\ding{56}} & \textcolor{scifiPurple}{\ding{56}} & & S & \textcolor{scifiPurple}{\ding{56}} & \textcolor{scifiBlue}{\ding{52}} & 50 \\
        OVO-Bench~\cite{liOVOBenchHowFar2025} & MC & \textcolor{scifiPurple}{\ding{56}} & \textcolor{scifiPurple}{\ding{56}} & \textcolor{scifiPurple}{\ding{56}} & & S & \textcolor{scifiPurple}{\ding{56}} & \textcolor{scifiPurple}{\ding{56}} & 2,814 \\
        OVO-Bench (FAR)~\cite{liOVOBenchHowFar2025} & C \& Q & \textcolor{scifiBlue}{\ding{52}} & \textcolor{scifiPurple}{\ding{56}} & \textcolor{scifiPurple}{\ding{56}} & & M & \textcolor{scifiBlue}{\ding{52}} & \textcolor{scifiPurple}{\ding{56}} & 1618 \\
        MMDuet~\cite{wang2024videollmknowsspeakenhancing} & OE & \textcolor{scifiBlue}{\ding{52}} & \textcolor{scifiPurple}{\ding{56}} & \textcolor{scifiPurple}{\ding{56}} & & M & \textcolor{scifiPurple}{\ding{56}} & \textcolor{scifiBlue}{\ding{52}} & 2000 \\
        \hline % 数据分组之间的细分隔线
        \textbf{\textit{ Ego Benchmark}} \\
        EgoPlan~\cite{chenEgoPlanBenchBenchmarkingMultimodal2024} & OE & \textcolor{scifiPurple}{\ding{56}} & \textcolor{scifiPurple}{\ding{56}} & \textcolor{scifiPurple}{\ding{56}} & & S & \textcolor{scifiPurple}{\ding{56}} & \textcolor{scifiPurple}{\ding{56}} & 5,000 \\
        EgoPlan2~\cite{qiuEgoPlanBench2BenchmarkMultimodal2024} & OE & \textcolor{scifiPurple}{\ding{56}} & \textcolor{scifiPurple}{\ding{56}} & \textcolor{scifiPurple}{\ding{56}} & & S & \textcolor{scifiPurple}{\ding{56}} & \textcolor{scifiPurple}{\ding{56}} & 1,300 \\
        EgoSDQES~\cite{eyzaguirreStreamingDetectionQueried2024} & Q-Match & \textcolor{scifiBlue}{\ding{52}} & \textcolor{scifiPurple}{\ding{56}} & \textcolor{scifiPurple}{\ding{56}} & & S & \textcolor{scifiBlue}{\ding{52}} & \textcolor{scifiBlue}{\ding{52}} & 3,971 \\
        \hline % 数据分组之间的细分隔线
        ESTP (Ours) & OE & \textcolor{scifiBlue}{\ding{52}} & \textcolor{scifiBlue}{\ding{52}} & \textcolor{scifiBlue}{\ding{52}} & & M & \textcolor{scifiBlue}{\ding{52}} & \textcolor{scifiBlue}{\ding{52}} & 2264 \\
        \specialrule{1pt}{0pt}{0pt} % 粗底线
    \end{tabular}
    }
    % 注意：下面的子标题需要调整跨列范围到 
    \captionsetup{width=0.96\linewidth}
    \captionof{table}{Comparison of datasets based on proactive and streaming criteria. This table summarizes datasets by Question Types (Open-Ended (OE); Multiple Choice (MC); Query Matching (Q-Match \& Q); and Count (C)), Proactive Types (Explicit (Exp.); Implicit (Imp.); and Contextual (Cont.)), and Just-in-Time (JIT) Responsiveness. Key JIT Responsiveness aspects include Answer Turn (Ans. Turn) (options: Single (S), Multi (M)), Precision (Is Prec.), and Timeliness. The notation '\# Ques.' denotes the number of questions.}
        \label{tal:t1}
    \end{minipage}%
    \vspace{-0.4cm}
\end{figure}

Unfortunately, existing evaluation frameworks~\cite{wang2024videollmknowsspeakenhancing,linStreamingBenchAssessingGap2024,liOVOBenchHowFar2025,zhangFlashVStreamMemoryBasedRealTime2024,eyzaguirreStreamingDetectionQueried2024} and streaming models~\cite{chenVideoLLMonlineOnlineVideo2024,wuVideoLLMMoDEfficientVideoLanguage2024} fall short in supporting or measuring the unified capabilities of proactive, just-in-time, and synchronized reasoning—and often struggle even with some individual aspects. 
Offline video benchmarks~\cite{fuVideoMMEFirstEverComprehensive2024,zhouMLVUComprehensiveBenchmark2024,mangalamEgoSchemaDiagnosticBenchmark2023, wang2024lvbenchextremelongvideo,ataallah2024infinibenchcomprehensivebenchmarklarge} evaluate video LLMs across diverse question types and scenarios, but their offline nature limits the assessment of the three core capabilities essential for online deployment. Recent efforts toward online and streaming benchmarks address this gap by introducing proactive tasks. Nevertheless, as shown in Tab ~\ref{tal:t1}, they often offer limited question diversity, lack contextual continuity across queries, and—more importantly—rarely evaluate just-in-time responsiveness or synchronized efficiency. As a result, current online video LLMs remain confined to narrow tasks such as narration or simple question answering, lacking the capacity for continuous, multi-turn understanding. More critically, as illustrated in Fig.~\ref{fig:exp2}, these models exhibit poor just-in-time behavior—often generating under-responsive or over-extended answers. Similarly, although recent efforts~\cite{wang2024videollmknowsspeakenhancing, qianDispiderEnablingVideo2025} have begun to address efficiency, they tend to focus solely on accelerating response generation—potentially at the cost of answer accuracy—while overlooking the need to balance perception and answering under synchronized constraints.

As a first step toward addressing these challenges, we introduce \textcolor{scifiPurple!50}{\textit{\textbf{a new Ego STreaming Proactive (ESTP) benchmark and evaluation framework}}}, specifically designed to capture the demands of the three key properties in streaming video. \textbf{For proactive coherence}, all question-answering tasks in the benchmark are proactive in nature: each question can only be answered based on future video streams within one or more specific time intervals. To reflect different levels realistic scenarios, we group them into three types: (1) explicit, grounded in clear visual cues; (2) implicit, requiring reasoning beyond surface observations; and (3) contextual, involving temporally linked questions that demand consistent multi-turn answers. We collect 2,264 questions spanning 14 task types—such as object localization, state change understanding, and intention prediction—across over 100 types of distinct scenarios, including kitchen activities, social interactions, and daily object manipulation. \textbf{For just-in-time responsiveness}, we emphasize the importance of response timing: each question are annotated an average of 3.96 valid answer intervals, and a prediction is considered valid only if it falls within the designated window. To assess this, we introduce ESTP-F1, a metric that integrates answer quality, response timing, and temporal precision. Additionally, 46\% of questions are contextually linked, requiring coherent responses based on prior questions—highlighting the need to continuously track the evolving stream from past to future and respond at the right moment. 
\textbf{For synchronized efficiency}, we not only evaluate time and memory efficiency and answering accuracy independently, but also assess accuracy under tightly synchronized perception and response—offering a comprehensive perspective on streaming video LLM evaluation. 

To address this novel task, \textcolor{scifiRed!50!black}{\textit{\textbf{we propose a comprehensive and novel technical pipeline—including a data engine, multi-stage training strategies, and a proactive dynamic compression technique—to enhance the streaming video LLMs. Specifically}}}, 
\begin{itemize}[leftmargin=*, itemsep=0pt]
\item \textbf{The data engine} automatically generates diverse, multi-turn questions and their corresponding answers to support the demands of continuous and proactive question answering. This involves a three-stage generation pipeline covering (1) one-to-one: using LVLMs to generate captions and extract initial question-answer pairs with a single temporal answer interval; (2) one-to-many: applying RAG to expand each answer into multiple valid intervals; and (3) many-to-many: composing coherent multi-turn questions from related QA pairs. 
\item \textbf{The multi-stage training strategy} is employed to progressively learn: (1) passive interval responsiveness, which provides a basic ability to trigger responses by distinguishing visually similar frames with different response labels, but often results in over-responsiveness even when the correct response interval; (2) Proactive just-in-time responsiveness and accurate answering, which trains the model to actively request high-resolution frames during uncertain timestamps, allowing it to use fine-grained visual details to pinpoint both the correct response moment and the accurate answer; (3) Coherence across multi-turn QA, which enables the model to maintain consistency by reasoning over prior QA history and current context, supporting contextual consistency answering. 
\item \textbf{The proactive dynamic compression} technique fully leverages the streaming nature by applying two levels of token compression based on response likelihood, including: (1) when the model anticipates a potential response, it proactively requests high-resolution inputs to improve the accuracy of perception and answering; (2) Otherwise, it applies a higher compression rate to past content to reduce token usage and improve efficiency; (3) Additionally, once a response is completed, the content preceding its timestamp is further compressed to free up resources without affecting future perception or answering.
\end{itemize}

% % 采用如下宏定义更方便
\newcommand{\cmark}{\ding{51}}
\newcommand{\xmark}{\textcolor{scifiPurple}{\ding{55}}}
\newcommand{\crossmark}{\ding{53}}

In summary, our contributions include: \textcolor{scifiGreen!70!black}{the novel Ego-Streaming Proactive (ESTP) task, distinguished by its three key properties}; \textcolor{scifiPurple!50}{the ESTP-Bench benchmark and the ESTP-F1 metric for robust evaluation of this task}; and \textcolor{scifiRed!50!black}{a comprehensive and novel technical pipeline, incorporating three key techniques, designed to address the ESTP task.} Our results demonstrate that the proposed model effectively overcomes the key challenges posed by this task. Moreover, it demonstrates superior performance by substantially exceeding multiple baselines in diverse online and offline benchmarks.

\section{Ego Streaming Proactivate Dataset \& Benchmark}

\subsection{Data Source and Annotation}
\textbf{Data Source} is validation set of Ego4D~\cite{graumanEgo4DWorld30002022,song2023egod} that includes raw annotations such as event narrations and steps for completing consistent goals. Following~\cite{linEgocentricVideoLanguagePretraining2022, chenVideoLLMonlineOnlineVideo2024}, we filtered out video with missing or uncertain annotations and converted annotations into a natural language format. This process yielded 890 videos, encompassing over 100 distinct scenes and a wide array of human activities, including indoor home environments (e.g., cooking, cleaning), workspaces (e.g., working at desk, labwork, baker), and public areas (e.g., grocery shopping). Furthermore, the videos exhibit rich dynamic diversity, ranging from periods of relative stillness (e.g., observing a static scene) to highly dynamic moments involving rapid manipulation tasks or active locomotion (e.g., cooking, walking).

\noindent\textbf{Annotation} process follows a two-step procedure. First, initial QA pairs are automatically generated with the assistance of MLLMs~\cite{yao2024minicpmvgpt4vlevelmllm, wang2024qwen2vlenhancingvisionlanguagemodels} and LLMs~\cite{deepseekai2025deepseekv3technicalreport}. Second, these automatically generated questions provided inspiration for annotators, aiding them in identifying valuable instances or formulating question ideas. To ensure diversity of questions, we annotate three proactive types:

\begin{figure}[h]
    \centering
    \begin{minipage}[h]{0.4\textwidth}
        % \vspace{0pt} % 关键：强制顶部对齐
        \includegraphics[width=\linewidth]{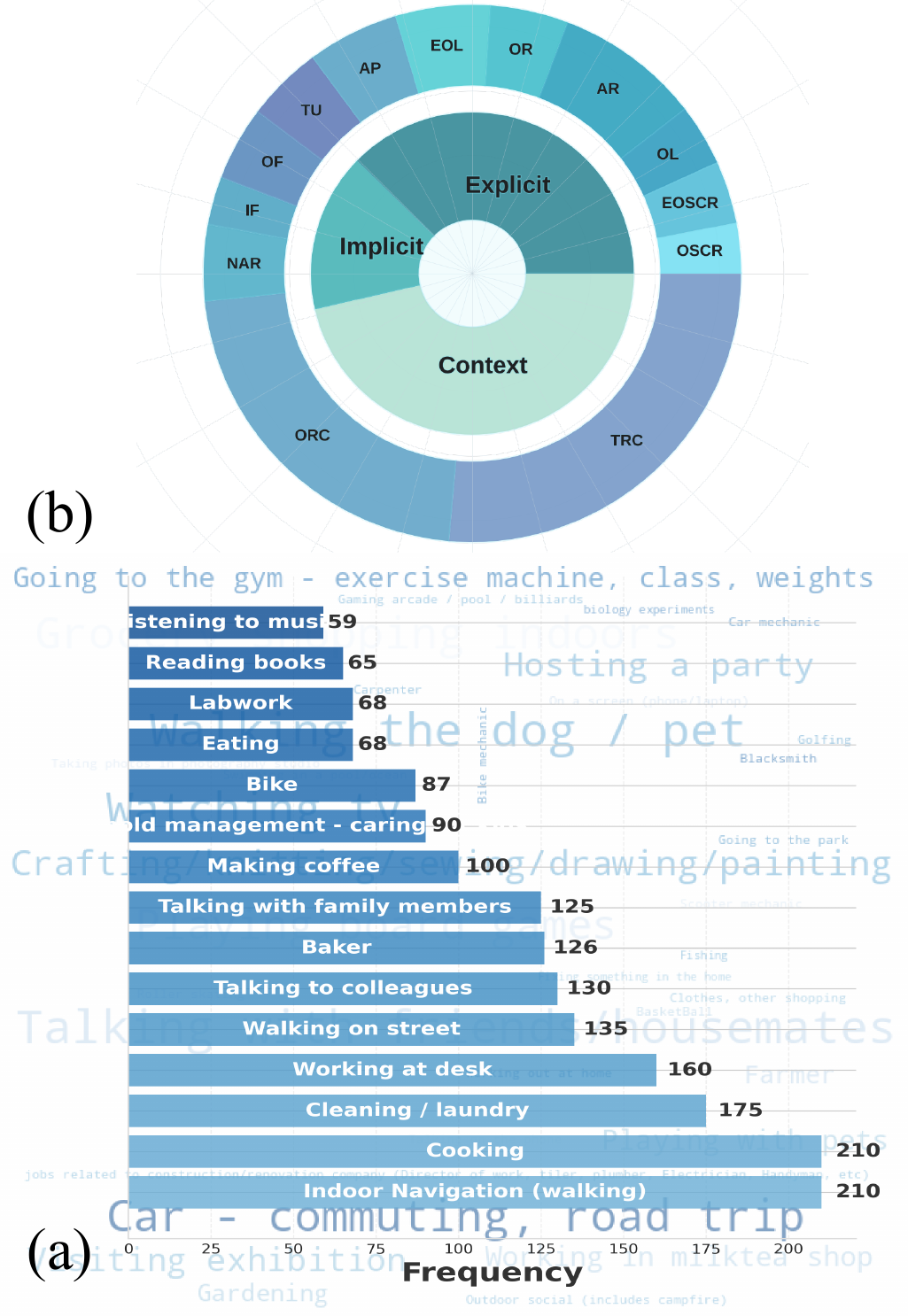}
        \caption{(a) Frequency of scenes or activities from which tasks and questions are derived. (b) Proportion of different proactive and question task types. }
        \label{fig:statis}
    \end{minipage}%
    \hspace{0.01\textwidth}
    \begin{minipage}[h]{0.54\textwidth}
        % \vspace{0pt} % 关键：强制顶部对齐
        % \small
        % \sloppy

\textbf{(1). \textit{Explicit Proactive Tasks}} are defined as those required to identify and respond to queries by directly leveraging and interpreting visual information present in the input. This category encompasses tasks where the relevant visual cues are explicitly referenced or are central to formulating a correct response. This category is comprised of eight distinct task types: Object Recognition (OR), Attribute Perception (AP), Text-Rich Understanding (TRU), Object Localization (OL), Object State Change (OSC), Ego Object Localization (EOL), Ego Object State Change (EOSC), and Action Recognition (AR).
\textbf{(2). \textit{Implicit Proactive Tasks}} are defined as those requiring inference and deeper scene understanding that goes beyond immediate, direct observation. This category is comprised of four distinct task types: Object Function Reasoning (OFR), Information Function Reasoning (IFR), Next Action Reasoning (NAR), and Task Understanding (TU).
\textbf{(3). \textit{Contextual Proactive Tasks}} are defined as those requiring the model to maintain awareness of dialogue history and visual coherence across temporally extended interactions. This category is comprised of two distinct task types: Object Relative Context (ORC) and Task Relative Context (TRC).
Fig.~\ref{fig:statis} illustrate dataset distribution.
    \end{minipage}
\end{figure}

To enable the evaluation of Just-in-Time Responsiveness and eliminate ambiguity in answer intervals, human annotators are required to mark clear time interval boundaries based on the completeness of objects within frames or the start/end of events. Simultaneously, questions with ambiguous references are filtered out (e.g., ``Remind me the location of the ceramic bowl.'' where multiple ceramic bowls might be present in different locations). Each sample's question, answer, and corresponding answer interval are verified by two annotators. This rigorous verification process resulted in a dataset of 2264 verified question-answer instances. Notably, every answer in the dataset is associated with precise temporal annotations. Statistical information regarding the annotated data is presented in Fig.~\ref{fig:statis}.

\subsection{Evaluation Metric in ESTP}
To comprehensively measure performance along three key evaluation aspects – answer quality, response timing, and temporal precision – we introduce the ESTP-F1 score. Here, we denote a ground truth item as $g_k$ with content $o_k$, and a prediction as $\hat{p}_l$ with content $\hat{o}_l$ and time $\hat{t}_l$. Evaluation components are defined for matched pairs $(\hat{p}_l, g_k)$, where $\hat{p}_l$ is a prediction that temporally matches $g_k$. For answer quality, an LLM~\cite{deepseekai2025deepseekv3technicalreport} is used to measure correctness, defined as a score $\mathcal{S}_\text{answer}(\hat{o}_l, o_k)$ for the predicted content $\hat{o}_l$ relative to the ground truth content $o_k$. For evaluating response timing, we go beyond simply considering recall (which inherently accounts for False Negatives (FN)) and employ a score $\mathcal{S}_\text{time}(\hat{t}_l, g_k)$ to more precisely measure timeliness. Furthermore, for temporal precision, we introduce precision, utilizing False Positives (FP) as a penalty term. These components contribute to the aggregated ground truth score $S(g_k)$, which replaces the traditional binary TP count. The ESTP-F1 score is computed as:
\begin{equation}
\text{ESTP-F1} = \frac{2 \times \sum_{k=1}^M S(g_k)}{2\sum_{k=1}^M S(g_k) + \text{\texttt{FP}} + \text{\texttt{FN}}},
\end{equation}
where $M$ is number of GT.
High answer quality (reflected by a high $S_\text{answer}$ score), effective response timeliness (characterized by high $S_\text{time
}$ for on-time responses and a low False Negative (FN) rate), and high precision (indicated by a low False Positive (FP) rate) collectively contribute to a high ESTP-F1 score.
More details are provided in the Appendix.

\section{Methodology: VideoLLM-EyeWO}

\begin{figure}[t]
    \centering
    \includegraphics[width=1.0\linewidth]{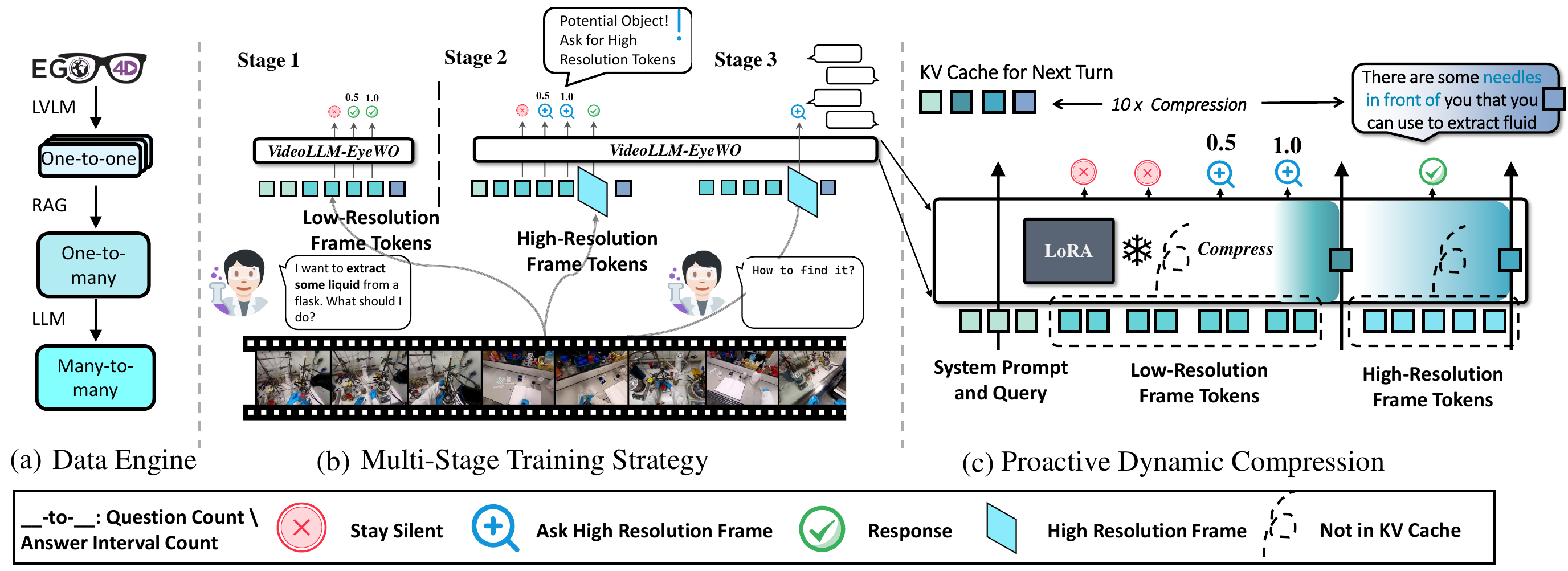}
    \caption{Overview of the proposed pipeline. The figure illustrates the three main components: (a) \textbf{Data Engine} (ESTP-Gen), which automatically generates diverse, multi-turn QA data through a three-stage pipeline. (b) \textbf{Multi-Stage Training Strategy} incrementally builds the model from basic responsiveness to proactive just-in-time accuracy, and ultimately to achieving multi-turn coherence, detailed in Section~\ref{sec:training}. (c) \textbf{Proactive Dynamic Compression}  detailed in Section~\ref{sec:compression}.}
    \label{fig:framework}
    \vspace{-0.2cm}
\end{figure}

In this section, we introduce a technical pipeline designed for the ESTP task.
For the data engine, utilizing the Ego4D~\cite{graumanEgo4DWorld30002022} training set and a three-stage generation pipeline as introduced in Sec.~\ref{sec:intro}, we generate 60K single-turn and 20K multi-turn questions, as shown in Fig.~\ref{fig:framework}.  Each generated instance includes questions, answers, and their corresponding valid answer intervals (named as ESTP-IT). See Appendix for data engine details.
Subsequently, we detail the problem definition and preliminary, the multi-stage training strategies, and the proactive dynamic compression technique in respective subsections.

\subsection{Problem Definition and Preliminary}

\textbf{Problem Definition.}
Given a streaming video input and a sequence of emerging queries $\mathcal{Q} = \{(q_i, t_{q_i})\}$, where $q_i$ is the query content and $t_{q_i}$ is the query timestamp. At each timestep $t$ following a query (i.e., $t > t_{q_i}$), the model must leverage its historical memory $H_t$ (including visual input history and past query-response interactions) , while concurrent observation $O_t$, to decide whether to perform a response action and generate corresponding content.
The model's decision-making process at time $t$ can be formulated as selecting the optimal action $A_t$ from a predefined set $A$:
\begin{equation}
A_t = \operatorname{argmax}_{a \in A} P_\theta(A_t = a \mid q_i, O_{t}, H_{t}).
\end{equation}
Here, $\theta$ represent model parameter, $A_t$ is the model's action at time $t$, and $A$ is the set of possible actions. Notably, while previous work typically considers an action space that only includes $a_{\text{silence}}$ (staying silent) and $a_{\text{response}}$ (executing a response and generating a reply), we expands this by including the action $a_{\text{ask\_high}}$ (requesting a high-resolution frame), as introduced in Sec.~\ref{sec:training} Stage-1.

\textbf{Preliminary.}
LIVE~\cite{chenVideoLLMonlineOnlineVideo2024} utilizes ground truth containing timestamps and applies cross-entropy supervision~\cite{vaswani2023attentionneed} to the model's action output at each timestep. 
Specifically, if the current time $t$ falls within a ground truth response region (denoted as $t \in \mathcal{T}_{\text{timestamp}}$), 
% where $\mathcal{T}_{\text{GT}}$ is defined by the set of all valid answer timestamps, 
the model is supervised to execute the response action ($a_{\text{response}}$) and generate a reply, incorporating a language modeling loss $\mathcal{L}_{\text{LM}}$~\cite{yenduri2023generativepretrainedtransformercomprehensive, devlin2019bertpretrainingdeepbidirectional, vaswani2023attentionneed}. Otherwise, it is supervised to remain silent ($a_{\text{continue}}$). This is formulated as:
\begin{equation}
\mathcal{L}(t) =
\begin{cases}
-\log P_\theta(a_{\text{response}} \mid q_i, O_{t}, H_{t}) + \omega \mathcal{L}_{\text{LM}}(t) & \text{if } t \in \mathcal{T}_{\text{timestamp}} \\
-\log P_\theta(a_{\text{continue}} \mid q_i, O_{t}, H_{t}) & \text{otherwise},
\end{cases},
\end{equation}
where, $\omega$ is a balancing coefficient weighting the language modeling objective.

\subsection{Multi-Stage Training Strategy}
\label{sec:training}
Following~\cite{chenVideoLLMonlineOnlineVideo2024}, VideoLLM-EyeWO utilizes the same network architecture and is trained using LoRA~\cite{hu2021loralowrankadaptationlarge}. However, the single-stage training and simple binary supervision strategy employed in~\cite{chenVideoLLMonlineOnlineVideo2024} can lead to training conflict due to the high similarity of adjacent frames in streaming inputs. This conflict necessitates a difficult trade-off between over-extended and under-responsive. To address these limitations, we employ a multi-stage training strategy designed to progressively endow the model with response capabilities. The following subsections detail each stage of this training strategy.

\textbf{Stage-1 : Passive Interval Responsivenes.}
To provide the basic ability for autonomous response triggering, we leverage the valid answer intervals within the ESTP-IT to achieve a progressive transition from silence to response. Specifically, if current time $t$ falling within a valid answer interval (where $\mathcal{T}_{\text{interval}}$ is defined as the set of all such intervals $[s_i, e_i]$), we apply a weighted degree of response supervision, rather than direct binary classification, using the following loss function:

\begin{equation}
\mathcal{L}(t) =
\begin{cases}
-\log \left( \colorbox{scifiBlue!10}{$f\left(\frac{\left | t-e  \right | }{\left | s-e  \right | }\right)$}\cdot P_\theta(a_{\text{response}} \mid q_i, O_{t}, H_{t}) \right) + \omega \mathcal{L}_{\text{LM}}(t) & \text{if } \colorbox{scifiBlue!10}{$\exists [s, e] \in \mathcal{T}_{\text{interval}}, t \in [s, e]$}  \\
-\log P_\theta(a_{\text{continue}} \mid q_i, O_{t}, H_{t}) & \text{otherwise}
\end{cases},
\label{equ:stage1}
\end{equation}
The function $f$ is a linear decrease map used as a weighting factor applied to the response probability loss.
The \colorbox{scifiBlue!10}{highlight} in Equ.~\ref{equ:stage1} is used to distinguish the components specific to this stage.

% It is typically a decreasing function of $\frac{|t-e|}{|s-e|}$ (for $t$ within the segment $[s, e]$), implementing the progressive transition by applying weaker supervision earlier in the segment ($t$ closer to $s$) and stronger supervision later ($t$ closer to $e$). 

\textbf{Stage-2 : Proactive just-in-time responsiveness and accurate answering.} % 更好的响应和更好的回复
To use fine-grained visual details to pinpoint both the correct response moment and the accurate answer, we train the model to actively request high-resolution frames during uncertain timestamps in this stage.
Specifically, we first introduce a third predefined action $a_{\text{ask\_high}}$. When the model executes this action at time $t$, it triggers the acquisition of the high-resolution frame $O^h_t$ corresponding to the current observation $O_t$ using the following loss function for training: $\mathcal{L}_\text{ask\_high}(t)$:
\begin{equation}
\mathcal{L}_\text{ask\_high}(t) =
\begin{cases}
-\log \left( f\left(\frac{\left | t-e \right | }{\left | s-e  \right | }\right)\cdot P_\theta(\colorbox{scifiBlue!10}{$a_{\text{ask\_high }}$} \mid q_i, O_{t}, H_{t}) \right) & \text{if } \colorbox{scifiBlue!10}{$t \in \mathcal{T}_{\text{uncertain}}$} \\
-\log P_\theta(a_{\text{continue}} \mid q_i, O_{t}, H_{t}) & \text{otherwise},
\end{cases},
\end{equation}
where $\mathcal{T}_{\text{uncertain}}$ denotes the set of the model's uncertain (see more detail in Appendix D Stage-2 Input). We use this loss to enable the model to acquire the ability to request high-resolution frames, and then based on the more detailed information, determine whether it is the correct time to respond and provide a more accurate answer, using the following loss:
\begin{equation}
\mathcal{L}_{\text{determine}}(t) =
\begin{cases}
-\log  P_\theta(a_{\text{response}} \mid q_i, O_{t}, H_{t}, \colorbox{scifiBlue!10}{$O^h_{t}$}) + \omega \mathcal{L}_{\text{LM}}(t) & \text{if } t \in \mathcal{T}_{\text{timestamp}} \\
-\log P_\theta(a_{\text{continue}} \mid q_i, O_{t}, H_{t}, \colorbox{scifiBlue!10}{$O^h_{t}$}) & \text{otherwise}
\end{cases},
\end{equation}
where $O^h_t$ represents the high-resolution frame acquired at time $t$.
The overall loss function at timestep $t$ is the sum of the two components:
\begin{equation}
\mathcal{L}(t) = \mathcal{L}_\text{ask\_high}(t) + \mathcal{L}_{\text{determine}}(t)
\end{equation}

See appendix for detailed uncertain timestamps $\mathcal{T}_{\text{uncertain}}$ identified.

\textbf{Stage-3 : Coherence across multi-turn QA.}
Building upon the model's acquired proactive and timely response capabilities, we introduce a separate training stage.
Specifically, this stage involves training solely on multi-turn question, with the aim of further improving its contextual understanding while preserving its timely responsiveness.

\subsection{Proactive Dynamic Compression Mechanism}
\label{sec:compression}
In order to ensure memory efficiency as the number of incoming frames grows over time, we propose the Proactive Dynamic Compression Mechanism, which applies two levels of token compression and employs a uniform compression method, detailed respectively in the following two subsections.

\textbf{Two-Level Compression.}
In contrast to fixed compression rates~\cite{liuVideoXLProReconstructiveToken2025,qianStreamingLongVideo2024, chenLLaVoltaEfficientMultimodal2024} and steps~\cite{shuVideoXLExtraLongVision2024, ryooXGenMMVidBLIP3VideoYou2024,yeVoCoLLaMAVisionCompression2024,zhangSoaring4K400K2024}, our mechanism leverages the streaming nature to allow the model to proactively determine both when to compress and which compression level to apply. 
Regarding the timing of compression, after the model generates a response, the preceding visual input and the response content itself form a natural segment or processing unit. Simultaneously, lower compression rates are applied to question-relevant content such as high-resolution frames, while higher rates are applied to other content, with these decisions proactively made by the model. Specifically, after a response, a fixed number of compression tokens (e.g., 1) are used to compress the preceding content, absorbing information from potentially many low-resolution frames or a single high-resolution frame. This approach naturally achieves a high compression rate for redundant parts of the past content, resulting in an average token usage of only about one-tenth of the original sequence.

\textbf{Uniform Compression Method.}
For achieving two-level compression, we employ a Uniform Compression Method. Specifically, unlike methods using additional compression modules~\cite{qianStreamingLongVideo2024, qianDispiderEnablingVideo2025}, we insert a special compression token ($\langle\mathrm{ct}\rangle$) after segments of original input, namely after single high-resolution frames, after multiple low-resolution frames, and after answer. This token is initialized using the text embedding of the ``<EOS>'' token. Leveraging the properties of the causal self-attention mechanism, this token prompts the LLM to compress the information from the preceding segment into a compact representation stored in the KV cache. 

During training, inspired by~\cite{shuVideoXLExtraLongVision2024}, the LLM is trained to process response turns sequentially. A response turn refers to a turn of interaction, typically a comprising visual input and a model's response. Training for the Proactive Dynamic Compression Mechanism, including the integration of high-resolution frame requests, commences in Stage 2 of our multi-stage training strategy to ensure manageable training memory overhead.
\section{Experiment}
\label{sec:exp}

\definecolor{techblue}{RGB}{0, 102, 204}
\definecolor{techgray}{RGB}{240, 240, 240}
\definecolor{techdark}{RGB}{30, 30, 30}
\definecolor{techaccent}{RGB}{0, 153, 255}

\begin{table*}[t]
% \label{expESTP}
\centering
% \caption{\textbf{Performance of Various MLLMs on ESTP Bench} }
% \renewcommand{\arraystretch}{1.5}

\huge
\resizebox{\textwidth}{!}{
\begin{tabular}{l| ccccccccc |ccccc |ccc |c}
\specialrule{3pt}{0pt}{0pt}  % 2pt 粗的横线
\multirow{2}{*}{\textbf{Model}}  & \multicolumn{9}{c|}{\textbf{\color{techdark}Explicit Proactive Task}} & \multicolumn{5}{c|}{\textbf{\color{techdark}Implicit Proactive Task}} & \multicolumn{3}{c|}{\textbf{\color{techdark}Contextual Q}} & \multirow{2}{*}{\textbf{\color{techdark}Overall}}  \\
\cmidrule(lr){2-10}  \cmidrule(lr){10-18}
 & OR & AP & TRU & OL & OSC & EOL & EOSC & AR & All & OFR & IFR & NAR & TU & All & ORC & TRC & All & \\
\midrule[1pt]

\rowcolor{techgray!50}
\multicolumn{19}{l}{\cellcolor{techgray!30}\textit{Offline MLLMs Response-in-Last}} \\
\rowcolor{techblue!5}
LLaVA-OneVision  & 7.2 & 11.5 & 4.9 & 10.0 & 4.9 & 6.9 & 5.6 & 3.2 & 6.8 & 3.8 & 6.3 & 11.6 & 29.8 & 12.9 & 10.8 & 5.7 & 8.2 & 8.7 \\
\rowcolor{techblue!5}
Qwen2-VL  & 11.7 & 8.1 & 14.9 & 10.5 & 1.7 & 8.9 & 10.6 & 6.0 & 9.0 & 10.2 & 4.4 & 26.5 & 49.5 & 22.6 & 13.3 & 9.4 & 11.3 & 13.3 \\
\rowcolor{techblue!10}
MiniCPM-V   & 12.3 & 12.6 & 10.7 & 13.7 & 8.6 & 7.5 & 11.9 & 5.5 & 10.4 & 11.8 & 9.2 & 36.0 & 55.3 & 28.1 & 32.6 & 25.4 & 29.0 & 18.1 \\
\rowcolor{techblue!5}
LLaVA-NeXT-Video  & 8.3 & 9.4 & 7.4 & 10.2 & 7.8 & 7.4 & 10.3 & 5.6 & 8.3 & 6.4 & 6.7 & 21.1 & 45.9 & 20.0 & 10.1 & 9.8 & 9.9 & 11.9 \\
\rowcolor{techblue!5}
InternVL-V2 & 9.3 & 14.6 & 9.5 & 10.6 & 1.7 & 6.3 & 3.0 & 3.6 & 7.3 & 3.3 & 9.2 & 15.5 & 28.2 & 14.0 & 16.9 & 15.6 & 16.2 & 10.5 \\

\midrule[0.8pt]

\rowcolor{techgray!30}
\multicolumn{19}{l}{\cellcolor{techgray!30}\textit{VLMs for Streaming Detection}} \\
\rowcolor{techblue!5}
CLIP  & 7.3 & 9.5 & 7.4 & 8.5 & 1.8 & 4.7 & 2.2 & 2.7 & 5.5 & 2.8 & 5.2 & 51.3 & 29.3 & 22.2 & 4.6 & 3.8 & 4.2 & 10.1 \\
\rowcolor{techblue!5}
LaViLa   & 8.4 & 10.7 & 9.0 & 9.1 & 3.1 & 5.4 & 3.6 & 4.3 & 6.7 & 7.8 & 10.0 & 56.2 & 34.4 & 27.1 & 9.4 & 28.9 & 19.2 & 14.3 \\
\rowcolor{techblue!10}
EgoVLP  & 10.5 & 11.0 & 8.7 & 8.5 & 5.5 & 5.6 & 5.3 & 4.4 & 7.4 & 6.2 & 10.7 & 58.4 & 48.3 & 30.9  & 8.0 & 25.3 & 16.6 & 15.5 \\
\midrule[0.8pt]

\rowcolor{techgray!30}
\multicolumn{19}{l}{\cellcolor{techgray!30}\textit{Offline MLLMs Polling Strategy}} \\
\rowcolor{techblue!5}
LLaVA-OneVision  & 8.3 & 8.8 & 22.8 & 25.4 & 13.5 & 9.8 & 9.6 & 10.3 & 13.6 & 20.3 & 20.9 & 35.9 & 49.9 & 31.8 & 14.6 & 1.9 & 8.2 & 18.0 \\
\rowcolor{techblue!5}
Qwen2-VL  & 13.7 & 13.5 & 15.4 & 29.5 & 8.0 & 15.4 & 16.6 & 10.9 & 15.4 & 17.8 & 19.8 & 56.4 & 63.1 & 39.3 & 13.0 & 7.7 & 10.4 & 21.3 \\
\rowcolor{techblue!10}
MiniCPM-V  & 14.9 & 16.8 & 17.1 & 26.8 & 7.7 & 12.9 & 12.5 & 13.1 & 15.2 & 15.9 & 21.0 & 46.8 & 62.2 & 36.5 & 24.3 & 28.9 & 26.6 & 22.9 \\
\rowcolor{techblue!5}
LLaVA-NeXT-Video & 15.6 & 14.6 & 21.9 & 26.8 & 12.8 & 14.2 & 13.5 & 12.3 & 16.5 & 18.6 & 23.2 & 44.9 & 51.6 & 34.6 & 19.9 & 7.7 & 13.8 & 21.3 \\
\rowcolor{techblue!5}
InternVL-V2  & 11.3 & 5.9 & 7.0 & 10.1 & 0.7 & 2.7 & 5.2 & 2.2 & 5.6 & 8.3 & 2.9 & 4.3 & 11.2 & 6.7 & 6.1 & 5.3 & 5.7 & 5.9 \\
\midrule[0.8pt]

\rowcolor{techgray!30}
\multicolumn{19}{l}{\cellcolor{techgray!30}\textit{Online MLLMs}} \\
\rowcolor{techblue!5}
LIVE(threshold=0.8) & 9.7 & 11.0 & 7.4 & 10.8 & 1.9 & 6.0 & 3.6 & 5.6 & 7.0 & 4.2 & 7.4 & 12.9 & 12.8 & 9.3 & 19.6 & 13.8 & 11.8 & 9.1 \\
\rowcolor{techblue!5}
LIVE(threshold=0.9)  & 11.2 & 13.9 & 7.9 & 13.2 & 5.6 & 9.4 & 6.0 & 8.9 & 9.5 & 5.8 & 8.9 & 41.0 & 46.7 & 25.6 & 11.3 & 26.5 & 18.9 & 15.5 \\
\rowcolor{techblue!10}
MMDuet  & 7.2 & 10.3 & 17.6 & 10.2 & 4.2 & 6.1 & 8.8 & 8.5 & 9.1 & 10.0 & 7.7 & 50.1 & 69.1 & 34.2 & 17.4 & 23.1 & 20.3 &  17.8 \\
\rowcolor{techblue!20}
\textbf{VideoLLM-EyeWO(Ours)} & \textbf{26.6} & \textbf{26.6} & \textbf{25.1} & \textbf{26.8} & \textbf{19.8} & \textbf{22.3} & \textbf{20.8} & \textbf{20.7} & \textbf{23.6} & \textbf{24.8} & \textbf{31.0} & \textbf{75.3} & \textbf{78.7} & \textbf{52.5} & \textbf{39.5} & \textbf{47.8} & \textbf{43.6} & \textbf{34.7}\\
\specialrule{3pt}{0pt}{0pt} % 2pt 粗的横线
\end{tabular}
}

\vspace{-0.2cm}
% \footnotesize
\caption{Experimental results of various models evaluated on the ESTP-Bench. We present performance across Explicit Proactive, Implicit Proactive, and Contextual Question task types, as well as the Overall score, for Offline MLLMs (Response-in-Last and Polling Strategy), VLMs for streaming detection, and Online MLLMs. Deep blue highlights the best overall performance, while blue indicates the best performance within each model category and evaluation setting group.}
\label{tab:ESTP_result}
\vspace{-0.5cm}
\end{table*}
% \input{table/ESTP_Bench_Recall}

% In this section, we first present comprehensive experiments and in-depth analyses of ESTP-Bench, and subsequently, we conduct evaluation and ablation studies on Videollm-EyeWO.

\subsection{Baseline and Evaluation Settings}

We evaluate three categories of models in this study: Offline MLLMs, VLMs, and Online MLLMs.

For \textbf{Offline MLLMs} we selected representative models from different open-source MLLM families, including LLaVA-OneVision~\cite{li2024llavaonevisioneasyvisualtask}, Qwen2-VL~\cite{wang2024qwen2vlenhancingvisionlanguagemodels}, MiniCPM-V~\cite{yao2024minicpmvgpt4vlevelmllm}, LLaVA-NeXT-Video~\cite{li2024llavanextinterleavetacklingmultiimagevideo}, and InternVL-V2~\cite{chenHowFarAre2024}. As offline MLLMs lack inherent proactive response capability, following previous studies \cite{linStreamingBenchAssessingGap2024,liOVOBenchHowFar2025,wang2024videollmknowsspeakenhancing,chenVideoLLMonlineOnlineVideo2024}, we employed two evaluation settings: (1) \textit{Response-in-Last}: The model processes the complete video and is tasked with generating textual reply with timestamps. (2) \textit{Polling Strategy}: The model is periodically queried at fixed time intervals. If the model indicates readiness, it is then prompted to generate the answer. 
%Considering Synchronized Efficiency, we set the querying frequency to $0.175$ queries per second and selected models around the 7B/8B scale. This frequency was chosen to match the model's response efficiency, as shown in Fig.~\ref{fig:exp}. 
Specific details regarding the prompts and hyperparameter used in these settings are provided in Appendix. 

Regarding \textbf{VLMs}, following the approach of SDQES \cite{eyzaguirreStreamingDetectionQueried2024}, we selected CLIP~\cite{radford2021learningtransferablevisualmodels}, LaViLa~\cite{zhao2022learningvideorepresentationslarge}, and EgoVLP~\cite{linEgocentricVideoLanguagePretraining2022} for evaluation. These models were evaluated by computing the similarity between each frame and the query, using $0.5$ as the threshold to determine responsiveness. Notably, as these models cannot generate open-ended replies, their reply score is set to $0$. % Replace cite_SDQES with your actual citation key

For \textbf{Online MLLMs}, we selected VideoLLM-Online~\cite{chenVideoLLMonlineOnlineVideo2024} and MMDuet~\cite{wang2024videollmknowsspeakenhancing}, which provide open-source weights and streaming inference code, for evaluation. For VideoLLM-Online, we experimented with different thresholds to assess its performance variations.

\subsection{Benchmarking in ESTP-Bench}

\textbf{Comparative Analysis of Baseline Models.} Tab.~\ref{tab:ESTP_result} shows the performance of different models across three proactive types and fourteen task types under various evaluation settings, the experimental results consistently demonstrate that ESTP tasks pose significant challenges for all current types of models. Analysis revealed variations across model categories, with certain models exhibiting stronger capabilities within their respective groups (e.g., MiniCPM-V~\cite{yao2024minicpmvgpt4vlevelmllm} and QwenVL-2~\cite{wang2024qwen2vlenhancingvisionlanguagemodels} among offline MLLMs aligning with previous work~\cite{chengVidEgoThinkAssessingEgocentric2024}, and temporal VLMs like LaViLa~\cite{zhao2022learningvideorepresentationslarge} and EgoVLP~\cite{linEgocentricVideoLanguagePretraining2022} outperforming spatially-focused models like CLIP~\cite{radford2021learningtransferablevisualmodels}). Furthermore, the evaluation strategy significantly impacts performance. Specifically, offline MLLMs showed a notable disparity, performing on average better under the Polling Strategy compared to the Response-in-Last strategy, with improvements up to 5.4\%. This highlights the effectiveness of ESTP-Bench in evaluating models from a timeliness perspective and underscores the limitations in temporal grounding of existing offline models.

\textbf{Performance of VideoLLM-EyeWO Against Baselines.} As presented in Tab.~\ref{tab:ESTP_result}, our proposed model achieved significant performance improvements across all proactive tasks. Compared to the baseline videoLLM-Online~\cite{chenVideoLLMonlineOnlineVideo2024}, our model demonstrated an improvement of +19.2\%. Furthermore, it outperformed the best-performing model, MiniCPM-V~\cite{yao2024minicpmvgpt4vlevelmllm}(using the Polling strategy), by +11.8\%. 

\subsection{In-Depth Analyses in ESTP-Bench}
% \begin{figure}[t]
%     \centering
%     \includegraphics[width=1.0\linewidth]{fig/屏幕截图 2025-05-15 055648.png}
%     \caption{Analysis of challenges and model performance on the ESTP-Bench. (a) Average performance and Ground Truth interval proportion across 14 tasks, illustrating challenges with coherent and contextual questions. (b) Recall-Precision trade-off for different models and evaluation settings, highlighting the difficulty in responding only when necessary. (c) Action Per Second (APS) versus ESTP Performance Score for various models and polling strategies, measured on an A40 GPU, demonstrating synchronization efficiency challenges.}
%     \label{fig:exp}
% \end{figure}

\begin{figure}[t]
    \centering
    \begin{minipage}[t]{0.32\textwidth} % Adjusted width, using [t] for top alignment
        \centering
        \includegraphics[width=0.98\linewidth]{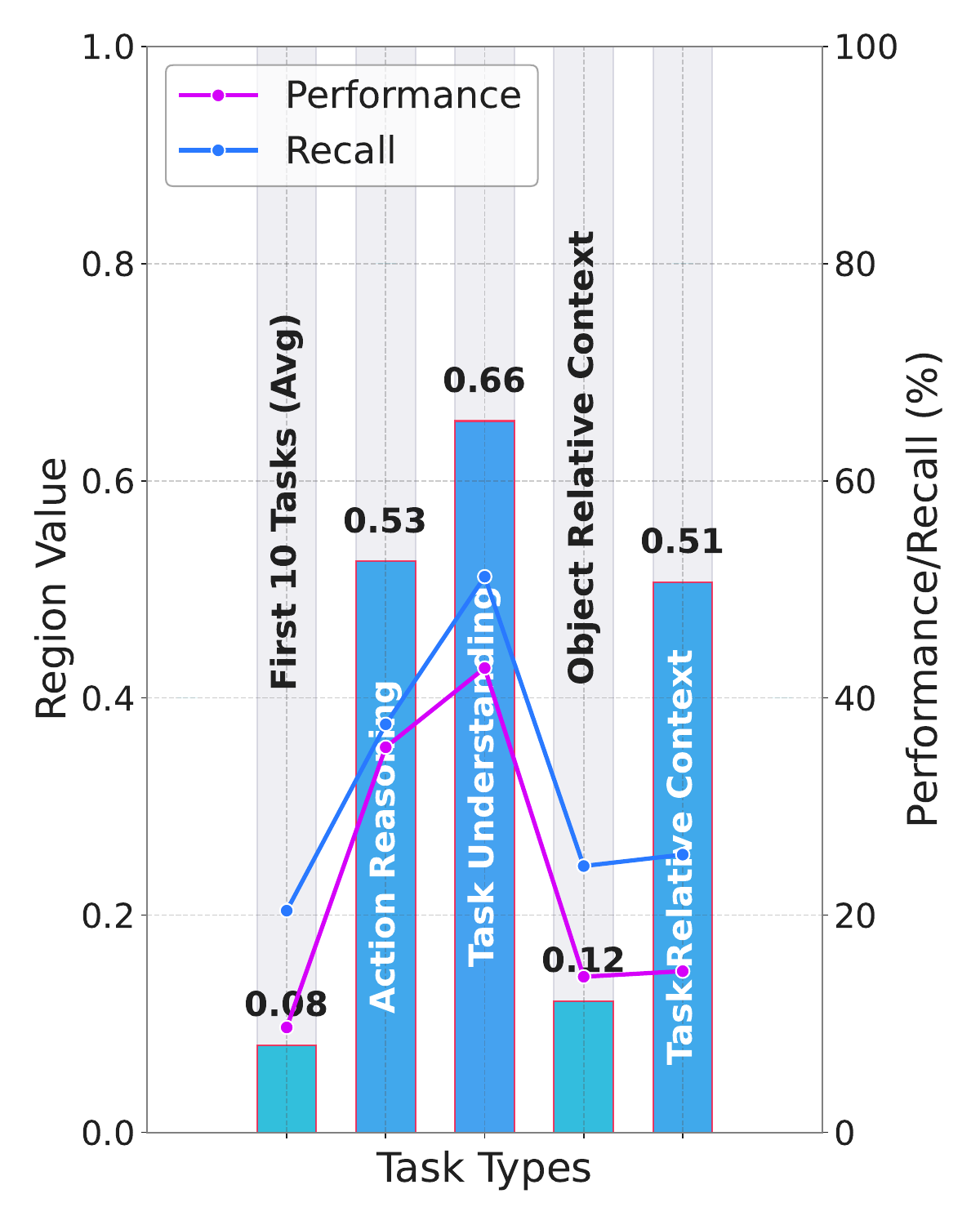}
        \caption{Average performance and Ground Truth interval proportion across 14 tasks, illustrating challenges with coherent and contextual questions.}
        \label{fig:exp}
    \end{minipage}%
    \hfill % Adds flexible horizontal space
    \begin{minipage}[t]{0.32\textwidth} % Adjusted width
        \centering
        \includegraphics[width=\linewidth]{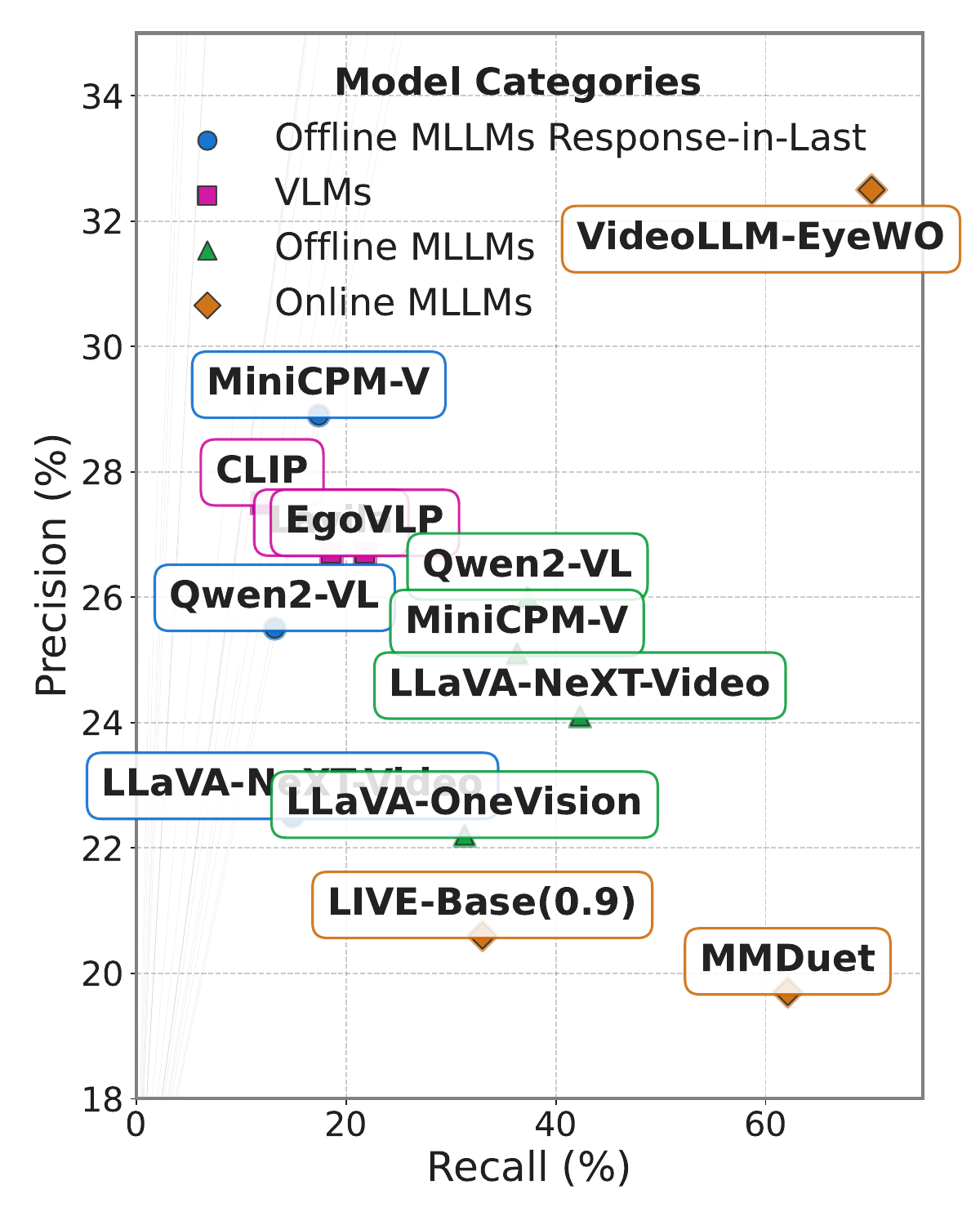}
        \caption{Recall-Precision trade-off for different models and evaluation settings, highlighting the difficulty in responding only when necessary.}
        \label{fig:exp2}
    \end{minipage}%
    \hfill % Adds flexible horizontal space
    \begin{minipage}[t]{0.32\textwidth} % Adjusted width
        \centering
        \includegraphics[width=\linewidth]{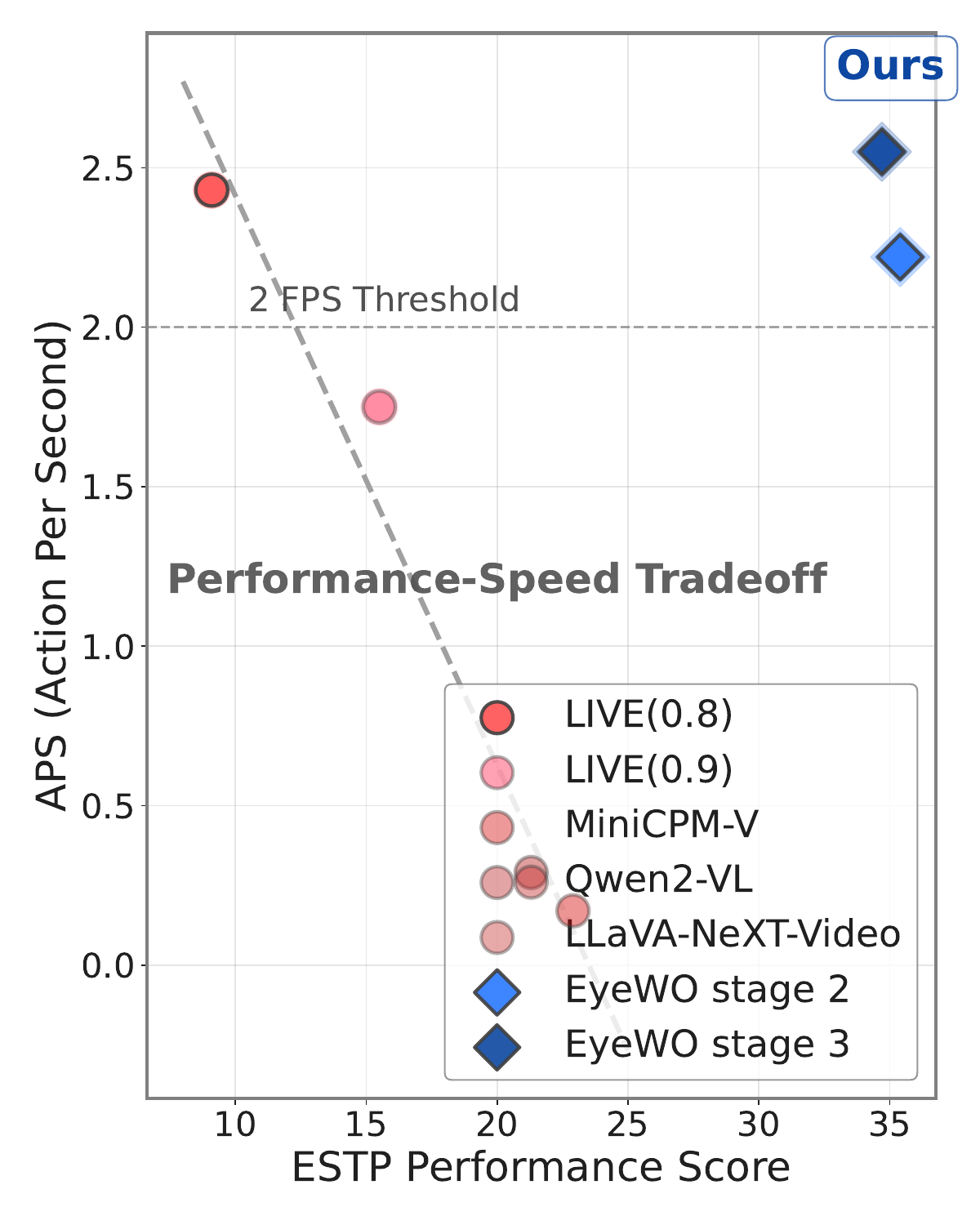}
        \caption{Action Per Second versus ESTP Score for various models, measured on an A40 GPU, demonstrating synchronization efficiency challenges.}
        \label{fig:exp3}
    \end{minipage}
    \vspace{-0.5cm}
\end{figure}

% \textbf{连贯的，上下文问题对于当前模型极具挑战：}Fig ~\ref{fig:task_perf_avg}绘制了14个任务下不同模型的平均性能（图中紫线），【NAR】Next Action Reasoning 和【TU】Task Understanding显著高于其他任务。然后，通过可视化这两个任务的GT区间占输入视频的比例显著高于其他任务，due to （思考一下这么表达，就是这些标注来自于GoalStep的原始标注，关于将一个具有一致目标的连续动作进行拆解，所以GT相关区域比例高），从而带来了较高的Recall。然而，我们发现，对于来自相同原始标注的、有着高比例的GT相关区域的【TRC】Task Relative Context任务的Recall和性能却显著降低。这表明proactive 连贯性对于现有模型极具挑战。
% \textbf{模型难以在必要时响应：} Fig ~\ref{fig:task_r_p}展示了不同评估设置下不同模型的recall和xxx-precision，我们发现在除去部分性能较低的模型，其他模型的recall和precision之间呈现负相关性，例如MMDuet具备极高的recall但是以较低的精度为代价，这表明现有模型难以进行主动的、精确的response
% \textbf{同步效率对于现有模型存在挑战：} Fig ~\ref{fig:task_r_p}LIVE（0.9）和不同offline MLLMs的轮询策略的每秒做出对于action的判断次数。结果表明，轮询设置不适合ESTP任务由于较低的效率，而LIVE虽然接近2FPS每秒（现有videollm最高的处理视频的采用频率），但是性能较差。
\textbf{Challenges with Coherent and Contextual Questions:}
Fig.~\ref{fig:exp} illustrates the average performance of different models across 14 tasks. (NAR) and (TU) exhibit significantly higher performance compared to other tasks. Upon visualizing the proportion of valid answer intervals relative to the input video duration for these two tasks, we observe that this proportion is substantially higher than for other tasks. This is attributed to these annotations originating from the raw GoalStep~\cite{song2023egod} labels, which involve segmenting continuous actions towards a consistent goal, thereby leading to a larger proportion of valid answer interval within the video and consequently, higher Recall. Conversely, for the (TRC) task, which also derives from the same original annotations and possesses a high proportion of valid answer interval, both Recall and overall performance significantly decrease. This marked performance drop underscores the significant challenge that proactive coherence and understanding contextual information pose for existing models.

\textbf{Difficulty in Responding Only When Necessary:}
Fig.~\ref{fig:exp2} presents the relationship between recall and precision for different models under various evaluation settings. We observe a prevalent negative correlation between recall and precision among most models. For instance, MMDuet achieves exceptionally high recall but at the expense of low precision. This trade-off indicates the struggle of existing models to provide proactive yet precise responses. 
%In contrast, our model achieved comprehensive superiority in both recall and precision. This significant lead demonstrates that our model effectively addresses the challenge of achieving timely and accurate responsiveness.

\textbf{Synchronization Efficiency Challenges:}
Fig.~\ref{fig:exp3} illustrates the inherent performance-speed tradeoff in ESTP tasks by plotting Action Per Second (APS) against Performance Score for various models. Existing methods often lie along a clear tradeoff curve, where higher performance is typically associated with lower APS, highlighting the difficulty in achieving both simultaneously. As seen for offline MLLMs using the Polling strategy, achieving high performance while maintaining sufficient speed for real-time synchronization remains challenging. Even approaches near the input frame rate (e.g., LIVE at $\sim$2 FPS) may demonstrate suboptimal performance. This underscores the significant challenges current models face in achieving both high task performance and effective synchronization with the dynamic video stream. 
% However, building on its precise responding, our model achieves significant efficiency gains. These gains enable it to process information faster than the input frame rate (2 FPS), effectively realizing a synchronous ``see and think'' capability and overcoming the typical performance-speed limitations.

\subsection{Evalutation of Videollm-EyeWO}

\textbf{Evaluating Zero-Shot Capability in Online/Offline Tasks.}
Table  ~\ref{tab:combined_detailed_performance_aligned_beautified} presents a performance comparison of our model against the baseline on both online and offline tasks. We selected VideoLLM-online~\cite{chenVideoLLMonlineOnlineVideo2024} as our baseline, given that it shares the same base model (LLaMA3~\cite{grattafiori2024llama3herdmodels} and SigLIP~\cite{zhai2023sigmoidlosslanguageimage}) and data source (Ego4D) as our own mode. For the online task, we utilize OvO-Bench~\cite{liOVOBenchHowFar2025} as a recognized benchmark. For the offline task, following~\cite{di2025rekv}, we evaluate our model on the multiple-choice subset of the QAEGO4D-test benchmark~\cite{barmannWhereDidLeave2022}. The `Online' setting involves posing questions as soon as the relevant answer segment appears, whereas the `Offline' setting involves questioning after the entire video has been presented. The experimental results demonstrate the generalization capability of our model across these distinct tasks.

\begin{table*}[ht]
    \centering
    \renewcommand{\arraystretch}{1.5} % 增大行间距
    % 使用 \small 调整字体大小以更好地适应 \textwidth
    \huge
    \resizebox{\textwidth}{!}{
    \begin{tabular}{l *{7}{c} *{4}{c} *{2}{c}}
        \specialrule{3pt}{0pt}{0pt} % 2pt 粗的横线
        % 第一层表头：任务分类 - Online Task 和 Offline Task
        \multirow{3}{*}{\textbf{Model}} & \multicolumn{11}{c}{\textbf{Online Task: OVO-Bench}} & \multicolumn{2}{c}{\textbf{Offline Task}} \\
        
        % 第二层表头：OVO-Bench 的子任务分类
        \cmidrule(lr){2-12} \cmidrule(lr){13-14}
        & \multicolumn{7}{c}{\textbf{Real-Time Perception}} & \multicolumn{4}{c}{\textbf{Backward Tracing}} & \multicolumn{2}{c}{$\textbf{QAEGO4D}_{\text{MC}}$} \\
        
        % 第三层表头：具体的指标名称
        \cmidrule(lr){2-8} \cmidrule(lr){9-12} \cmidrule(lr){13-14}
        & OCR & ACR & ATR & STU & FPD & OJR & \textbf{Avg.} & EPM & ASI & HLD & \textbf{Avg.} & Online & Offline \\
        \midrule
        
        VideoLLM-online & 8.05 & 23.85 & 12.07 & 14.04 & \textbf{45.54} & 21.20 & 20.79 & 22.22 & 18.80 & \textbf{12.18} & 17.73 & 29.80 & 30.20 \\
        \rowcolor{techblue!10}
        Ours (VideoLLM-EyeWO) & \textbf{24.16} & \textbf{27.52} & \textbf{31.89} & \textbf{32.58} & 44.55 & \textbf{35.87} & \textbf{32.76} & \textbf{39.06} & \textbf{38.51} & 6.45 & \textbf{28.00} & \textbf{36.20} & \textbf{33.00} \\
        \specialrule{3pt}{0pt}{0pt} % 2pt 粗的横线
    \end{tabular}
    }
    \caption{Detailed Performance Evaluation on OVO-Bench~\cite{liOVOBenchHowFar2025} and QAEGO4D~\cite{barmannWhereDidLeave2022} Tasks.}
    \label{tab:combined_detailed_performance_aligned_beautified}
\end{table*}

% \textbf{Evaluating VideoLLM-EyeWO in Other Baseline.}
% \input{table/livecc_eyewo}

\begin{figure}[ht]
% \vspace{-0.7cm}
    \begin{minipage}[h]{0.4\textwidth}
        \vspace{0pt} % 关键：强制顶部对齐
        % \small
        % \sloppy
\textbf{Evaluating Architecture Generalizability on Offline Tasks}
% 如Tab~\ref{table:COIN}所示，我们的模型在traditional temporal summarization and forecasting problems五个任务上取得了全面的提升，最最多能够提升+2.9%，这表明我们的模型结构能够有效泛化到其他Offline Tasks上。
As presented in Tab.~\ref{tab:COIN_result}, our model demonstrated comprehensive performance improvements on five tasks related to traditional temporal summarization and forecasting problems. The performance gain reached up to +2.8\%, which indicates that our proposed model architecture can effectively generalize to other offline tasks.
    \end{minipage}
    \hspace{0.01\textwidth}
    \centering
    \begin{minipage}[h]{0.5\textwidth}
        \vspace{0pt} % 关键：强制顶部对齐

\centering
\resizebox{\textwidth}{!}{
\begin{tabular}{c|ccccc}
% \specialrule{1.5pt}{0pt}{0pt} % 2pt 粗的横线
\toprule
\multirow{2}{*}{Method}   & \multicolumn{5}{c}{\textbf{COIN Benchmark}} \\

   & {Step} & {Task} & {Next} & {Proc} & {Proc+} \\
 
% \specialrule{1.5pt}{0pt}{0pt} % 2pt 粗的横线

% TimeSformer & $\times$ & 46.5 & 85.3 & 34.0 & 17.0 & 40.1 \\

% Paprika & $\times$ & 51.0 & 95.8 & 43.2 & - & - \\

% DistantSup & $\times$ & 54.1 & 90.0 & 39.0 & - & - \\

% VideoTF & $\times$ & 55.8 & 95.4 & 40.2 & 46.4 & - \\

% ProcedureVRL &$\times$ & 56.9 & 90.5 & 46.8 & - & - \\

% VideoTaskGraph & $\times$ & 57.2 & 91.2 & 42.7 & - & - \\

% \specialrule{1.5pt}{0pt}{0pt} % 2pt 粗的横线
\midrule
ClipBERT~\cite{lei2021moreclipbertvideoandlanguagelearning}  & 30.8 & 65.4 & - & - & - \\

VideoLLM-online-7B-v1~\cite{chenVideoLLMonlineOnlineVideo2024}  & 59.8 & 92.1 & 44.7 & 47.9 & 52.9 \\

\rowcolor{techblue!5}
VideoLLM-online-8B-v1+~\cite{chenVideoLLMonlineOnlineVideo2024}  & 63.1 & 92.6 & 49.0 & 49.7 & 53.6 \\

VideoLLM-MOD~\cite{wuVideoLLMMoDEfficientVideoLanguage2024}  & 63.4 & 92.7 & 49.8 & 49.8 & 53.3 \\

% StreamMind  & 63.7 & \underline{93.2} & 49.9 & 49.8 & 54.2 \\
\rowcolor{techblue!10}
Ours (LLaMa3~\cite{grattafiori2024llama3herdmodels,touvron2023llamaopenefficientfoundation}) & \underline{65.9} & \underline{92.7} & \underline{50.9} & \underline{50.8} & \underline{54.7} \\
\rowcolor{techblue!20}
Ours (LLaMa3.1~\cite{grattafiori2024llama3herdmodels,touvron2023llamaopenefficientfoundation})  & \textbf{66.0} & \textbf{93.3} & \textbf{51.5} & \textbf{51.1} & \textbf{55.5} \\
 \bottomrule
% \specialrule{1.5pt}{0pt}{0pt} % 2pt 粗的横线
\end{tabular}}
\captionsetup{width=0.96\linewidth}
    \captionof{table}{COIN~\cite{tangCOINLargescaleDataset2019} Benchmark Top-1 Accuracy comparison across different methods.}
\label{tab:COIN_result}

    \end{minipage}%
    
\end{figure}

\subsection{{Ablation Study of VideoLLM-EyeWO}}
\label{sec:ablation}

% \begin{table}[t]
%   \centering
%   \caption{Ablation study results on ESTP bench} % 表格标题 (Lowercase except for first word and proper nouns, NIPS style)
%   \label{tab:ablattion_two_level_header} % 表格标签
%   \begin{tabular}{lcccc} % 定义5列: 左对齐 (l), 居中 (c), 居中 (c), 居中 (c), 居中 (c)
%     \toprule % 表格顶部粗线
%     % 第一层表头
%     & \multicolumn{2}{c}{Single Question} & \multicolumn{2}{c}{Contextual Question} \\
%     \cmidrule(lr){2-3} \cmidrule(lr){4-5} % 第一层表头下的部分分隔线 (左右留白)
%     % 第二层表头
%     Method & Performance $\uparrow$ & KV Cache Size $\downarrow$ & Performance $\uparrow$ & KV Cache Size $\downarrow$ \\
%     \midrule % 表头和数据之间的中等粗细分隔线
%     \rowcolor{techblue!5}
%     LIVE & 12.9 & 9636.0 & 16.6 & 31199.5 \\
%     + ESTP-IT & 19.9 & 7859.1 & 23.1 & 28236.4 \\
%     + Stage-0 & 23.0 & 7988.2 & 21.8 & 17567.6 \\
%     \midrule % 在前面几行和应用新机制的行之间添加分隔线
%     \multicolumn{5}{c}{\textit{with increased proactive dynamic compression mechanism}} \\ % 跨列子标题 (斜体)
%     \midrule % 子标题下添加分隔线
%     \rowcolor{techblue!10}
%     + Stage-1 & \textbf{30.6} & 1182.8 & 34.7 & 3731.8 \\ % 加粗Single Question最高性能
%     \rowcolor{techblue!20} + Stage-2 & 29.1 & \textbf{942.0} & \textbf{38.4} & \textbf{3242.8} \\ % 加粗Contextual Question最高性能并整行着色
%     \bottomrule % 表格底部粗线
%   \end{tabular}
%   \label{tab:ab_result}
% \end{table}

\begin{table}[h]
  \centering
  \small
  \begin{tabular}{lcccc} % 定义5列: 左对齐 (l), 居中 (c), 居中 (c), 居中 (c), 居中 (c)
    \toprule % 表格顶部粗线
    % 第一层表头
    & \multicolumn{2}{c}{Single Question} & \multicolumn{2}{c}{Contextual Question} \\
    \cmidrule(lr){2-3} \cmidrule(lr){4-5} % 第一层表头下的部分分隔线 (左右留白)
    % 第二层表头
    Method & Performance $\uparrow$ & KV Cache Size $\downarrow$ & Performance $\uparrow$ & KV Cache Size $\downarrow$ \\
    \midrule % 表头和数据之间的中等粗细分隔线
    % \rowcolor{scifiBlue!5}
    LIVE & 14.9 & 9636.0 & 18.9 & 31199.5 \\
    + ESTP-IT & 22.0 & 7859.1 & 25.7 & 28236.4 \\
    \midrule % 表格顶部粗线
    Stage-0 & 24.9 & 7988.2 & 23.0 & 17567.6 \\
    % \midrule % 在前面几行和应用新机制的行之间添加分隔线
    \multicolumn{5}{c}{\textit{with increased proactive dynamic compression mechanism}} \\ % 跨列子标题 (斜体)
    % \midrule % 子标题下添加分隔线
    
    % \textcolor{gray}{+ Stage-1 ask low frame} & \textcolor{gray}{27.3} & \textcolor{gray}{623.1} & \textcolor{gray}{25.1}  &  \textcolor{gray}{1592.3} \\ % 加粗Single Question最高性能
    \rowcolor{techblue!10}
    + Stage-1 ask high frame & \textbf{34.0} & 1182.8 & 38.7 & 3731.8 \\ % 加粗Single Question最高性能
    \rowcolor{techblue!20} + Stage-2 & 33.2 & 942.0 & \textbf{43.6} & 3242.8 \\ % 加粗Contextual Question最高性能并整行着色
    \bottomrule % 表格底部粗线
  \end{tabular}
    \caption{Ablation study results on ESTP bench} % 表格标题 (Lowercase except for first word and proper nouns, NIPS style)
  \label{tab:ablattion_two_level_header} % 表格标签
\end{table}

% \input{table/coin_ablation}

% \textbf{Ablation and Efficiency Analysis}
Tab.~\ref{tab:ablattion_two_level_header} details the results of our ablation study on the ESTP benchmark:
\begin{enumerate}[leftmargin=*, itemsep=0pt]
\item (\textit{+ESTP-IT}) enhanced the LIVE baseline's performance on both Single and Contextual Question tasks, increasing it by +7.1 and +6.8 respectively, \textbf{thereby demonstrating the effectiveness of ESTP-IT}.
\item (\textit{Stage-0}) addressed the training conflicts stemming from simple binary supervision, enabling performance improvements without requiring any manual threshold tuning, \textbf{which demonstrates the model's acquisition of a basic ability to trigger responses}.
\item With the increased proactive dynamic compression mechanism, the model's KV cache consumption was significantly reduced, \textbf{requiring on average only about 0.11\% of the baseline}.
\item (\textit{+Stage-1}) significantly boosted Single Question performance to 34.0 and Contextual Question performance jumped to 38.7 by \textbf{incorporating the mechanism for actively requesting high-resolution frames for scrutiny alongside initial compression}.
\item (\textit{+Stage-2}) \textbf{further improved contextual coherence and refined compression}, enabling the model to achieve a gain of +4.9 on Contextual tasks, reaching 43.6. Simultaneously, the more accurate and efficient responses further reduced memory consumption to minimal levels.
\end{enumerate}

\section{Conclusion}
\vspace{-0.2mm}
We definite an novel AI assistant's task of proactive, synchronized question answering from ego-streaming video, targeting the key properties of proactive coherence, just-in-time responsiveness, and synchronized efficiency. Our contributions—the ESTP-Bench with its ESTP-F1 metric for evaluation, and a novel technical pipeline incorporating a data engine, multi-stage training, and proactive dynamic compression—enable our model to effectively tackle these properties. This approach outperforms multiple baselines across diverse online and offline benchmarks.

% \clearpage

\newpage

{
    \small
    \bibliographystyle{plain}
    \bibliography{main}
}

\end{document}